# Globally-scalable Automated Target Recognition (GATR)

Gary Chern, Austen Groener, Mike Harner, Tyler Kuhns, Andy Lam, Stephen O'Neill, and Mark Pritt
Lockheed Martin Space: Palo Alto, California; Valley Forge, Pennsylvania; and Rockville, Maryland

*Abstract*—GATR (Globally-scalable Automated Target Recognition) is a Lockheed Martin software system for real-time object detection and classification in satellite imagery on a worldwide basis. GATR uses GPU-accelerated deep learning software to quickly search large geographic regions. On a single GPU it processes imagery at a rate of over 16 km$^2$/sec (or more than 10 Mpixels/sec), and it requires only two hours to search the entire state of Pennsylvania for gas fracking wells. The search time scales linearly with the geographic area, and the processing rate scales linearly with the number of GPUs. GATR has a modular, cloud-based architecture that uses Maxar's GBDX platform and provides an ATR analytic as a service. Applications include broad area search, watch boxes for monitoring ports and airfields, and site characterization. ATR is performed by deep learning models including RetinaNet and Faster R-CNN. Results are presented for the detection of aircraft and fracking wells and show that the recalls exceed 90% even in geographic regions never seen before. GATR is extensible to new targets, such as cars and ships, and it also handles radar and infrared imagery.

## I. Background

*Automatic Target Recognition (ATR)* is the detection and recognition of targets, or objects of interest, in remotely sensed image data [1-4]. It has a long and rich history closely associated with the detection of enemy aircraft and bombers in radar data during World War II. Afterwards, ATR developed into the science of detecting objects in radar, optical, and infrared imagery from airborne and spaceborne sensors for the purposes of weapons targeting, reconnaissance, and navigation.

Closely associated with ATR is *image interpretation*, which is the analysis of objects and features in remotely sensed imagery [5]. Examples include aircraft, vehicles, ships, oil wells, deforested land, and construction sites. Applications include the monitoring of military equipment, economic forecasting, land management, archeology, and natural disaster response. Although image interpretation is performed manually by humans, there is room for automation. The application of deep learning to ATR shows promise for reducing the manual labor of image interpretation by performing tasks such as searching very large images or monitoring sites for change.

This paper introduces a Lockheed Martin software system called GATR that uses deep learning ATR algorithms to assist with the tasks of image interpretation and geospatial analysis. It applies these algorithms to find objects of interest anywhere in the world, perform broad area search, and monitor "watch boxes" over sites of interest.

In the field of deep learning, ATR is usually referred to as *object detection and classification*. It is the problem of finding objects in an image and classifying them into an object class (e.g., airplane, car, or ship) or type (e.g., Boeing 747 jet or C-130 cargo plane). This paper uses the term *ATR* synonymously with *object detection and classification*.

## II. Architecture

GATR, or Globally-scalable ATR, is a software system that uses deep learning ATR algorithms to assist with the process of image interpretation by searching large geographic regions for objects of interest. This is sometimes called *broad area search*. It also monitors sites for changes of interest. GATR has the following features:

- Global Scalability: It finds objects of interest anywhere in the world.
- Cloud Architecture: It is modular, cloud capable, and provides an ATR "analytic as a service".
- Speed: Graphical processing units provide acceleration.
- Accuracy: It uses state-of-the-art deep learning detection and classification algorithms.
- Extensibility: It can be trained on practically any type of object, facility, or feature, including aircraft, vehicles, ships, oil wells, and construction sites.

Fig. 1 shows the architecture of GATR. Satellite data flows to a data server, which supplies imagery to the training and detection algorithms of GATR on Amazon's AWS [6]. The satellite imagery is provided by Maxar Technologies' GBDX platform [7]. Detection results are saved in a geospatial database and sent to end users for analysis.

GATR uses two types of imagery from GBDX. WorldView-3 [8] is a high-resolution multispectral imaging satellite that provides imagery with a ground sample distance (GSD) of 0.3 meter. It is suitable for finding objects as small as cars. Sentinel-2 [9] is a low-resolution multispectral imaging satellite that provides imagery with a GSD of 10 meters. It is suitable for finding large objects and facilities such as oil and gas fracking wells. Its chief advantage over high-resolution satellites like WorldView-3 is its high revisit rate of 5-12 days. This makes it possible to collect sequences of satellite images for time-series characterization and monitoring of an object or facility [10].

Although GATR works primarily with multispectral imagery, it has also been tested with other types of imagery, including panchromatic, short wave infrared, and radar imagery such as Sentinel-1 [11].

---

The authors are listed alphabetically.

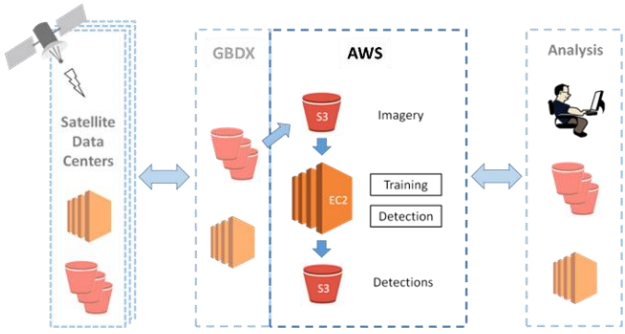

Fig. 1. Cloud architecture of GATR. Satellite imagery (left) flows to GBDX, which supplies imagery to the training and detection algorithms of GATR on AWS (center). Results are sent to databases and image analysts (right).

## III. DEEP LEARNING ATR

Signal processing techniques, such as correlation filters, feature extractors, and matching templates, have dominated the field of ATR for decades. Recently, *deep learning* techniques, driven by computer vision applications such as self-driving cars, have become very popular. Deep learning is a class of machine learning techniques that represent and manipulate data in large convolutional neural networks or CNNs [12,13]. These networks learn to recognize objects by means of supervised training on labeled image examples of the objects. Unlike conventional ATR methods, deep learning does not require the algorithm designer to engineer feature detectors. The networks themselves learn which features to detect, and how to detect them, as they train. These networks have achieved success in ATR and revolutionized computer vision and image understanding by combining CNNs with powerful graphical processing units (GPUs). Open source deep learning software libraries such as TensorFlow [14], PyTorch [15], and Keras [16] have helped to fuel continuing advances.

Early applications of deep learning to ATR limited themselves to the classification, but not detection, of image chips [6]. Current applications deal with the full ATR problem of detecting and classifying objects in large satellite images [13,17]. We have found that end-to-end deep learning algorithms like RetinaNet [18] and Faster R-CNN [19] are very effective if tuned to the characteristics of satellite imagery [17].

### A. Algorithm Training

Deep learning algorithms work in two stages. The first is an offline stage of supervised training, where the algorithm learns to recognize objects of interest as shown in Fig. 2. Object locations and times are compiled, and matching satellite images are retrieved. The objects, marked with labeled bounding boxes in the images, form the initial dataset. Next follows a labor-intensive step called dataset curation, in which human analysts correct the dataset. Objects obscured by cloud cover, for example, are removed, and unlabeled objects are labeled. The resulting dataset consists of images with labeled bounding boxes around the objects to be recognized. (The dataset must consist of real images and not just image "chips" of the objects.) The dataset is randomly split into two sets: one for training (typically 70-80% of the data) and the other for validation. The validation set is used to ensure that the algorithm does not "overfit" itself to the training data.

### B. Algorithm Inference

After the algorithm is trained, it is ready for operational use (called *inference*) as illustrated in Fig. 3. The algorithm monitors a bucket or folder of imagery, processing new images as they arrive and saving the detected objects to a database.

### C. Measuring Accuracy

The trained ATR algorithm is evaluated with a number of accuracy measures. After it is run in inference mode on the test (or validation) dataset, the true detections (*TP* or true positives), false detections (*FP* or false positives), and missed targets (*FN* or false negatives) are counted. The software then calculates the *recall*, *precision*, and $F_1$ *score* according to the following formulas:

$$P = \frac{TP}{TP+FP} \qquad R = \frac{TP}{TP+FN} \qquad F_1 = \frac{2PR}{P+R}$$

The recall is the same as probability of detection. The precision is the proportion of detections that are correct, and the $F_1$ score is the harmonic mean of the recall and precision.

The software generates a *precision-recall curve* by varying the detection threshold and plotting the precision and recall values. Just as the ROC curve from conventional ATR shows the trade-off between the probability of detection and the false alarm rate [1-4], the precision-recall curve shows the trade-off between the recall and the precision. One can adjust the detection threshold to increase the recall, but this comes at the cost of decreasing the precision.

The *mean average precision*, or *mAP*, is a standard measure of the detection and localization accuracy. It averages the precision over all recall values as a function of the intersection over union (IoU), which measures the degree of overlap between a predicted bounding box and the true bounding box that contains a detected object.

### D. Results on Aircraft Detection

The RetinaNet [18] and Faster R-CNN [19] models were trained on the 719 images of "Passenger and Cargo Aircraft" in the xView dataset [20,21]. Examples of these aircraft are shown in Fig. 4. They were supplemented with 362 military airplanes curated from 20 WorldView-3 images (Fig. 5). Results of the inference stage are shown in Figs. 6 and 7.

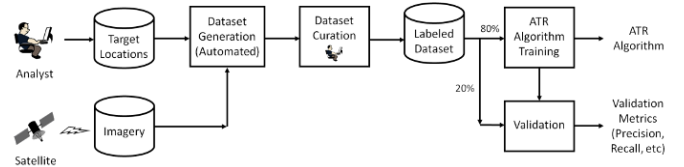

Fig. 2. Training stage. Analysts identify the locations of objects of interest and retrieve satellite images of these objects. The most labor intensive step is the manual labeling of the objects ("Dataset Curation").

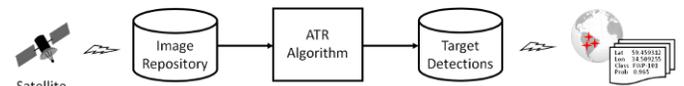

Fig. 3. Operational (or inference) stage. Once the ATR algorithm is trained, it is ready for the operational mission of finding objects in imagery.

The accuracy measures and precision-recall curves are shown in Fig. 8. For Faster R-CNN, the recall was 0.92 and the precision 0.96. The $F_1$ score was 0.94 and the mAP was 0.91.

*E. Results on Fracking Well Detection*

The algorithms were next trained on 1142 fracking wells in Pennsylvania. These wells were curated in WorldView-3 and Sentinel-2 imagery as two different datasets to determine the effect of GSD on ATR accuracy. Fig. 9 shows how a fracking well appears in each image. The results are shown in Fig. 10. For Faster R-CNN on WorldView-3, the recall was 0.94 and the precision was 0.90. The $F_1$ score was 0.92 and the mAP was 0.89. For lower-res Sentinel-2, the $F_1$ and recall were 0.90 and 0.88, respectively—only 6% lower than for WorldView-3.

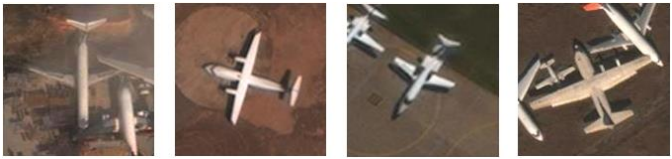

Fig. 4. Examples of "Passenger and Cargo Aircraft" in the xView dataset.

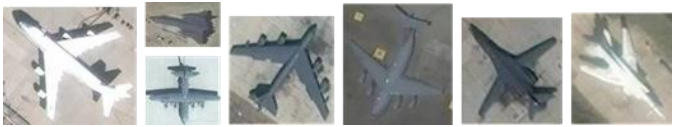

Fig. 5. Examples of military aircraft collected and curated to supplement the xView aircraft dataset. (Images: DigitalGlobe, Inc.)

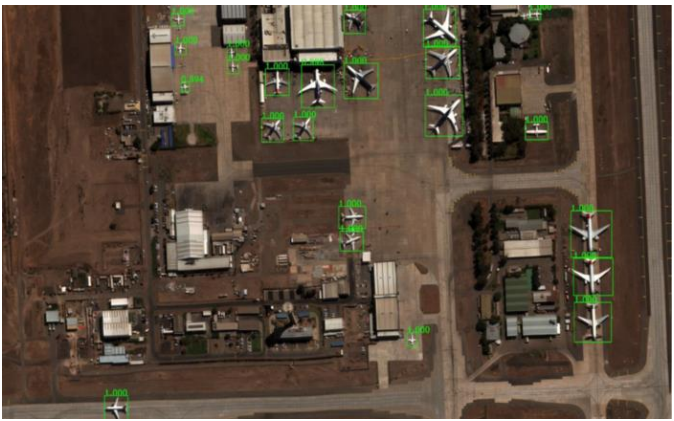

Fig. 6. Aircraft detection results (green boxes) at a commercial airport.

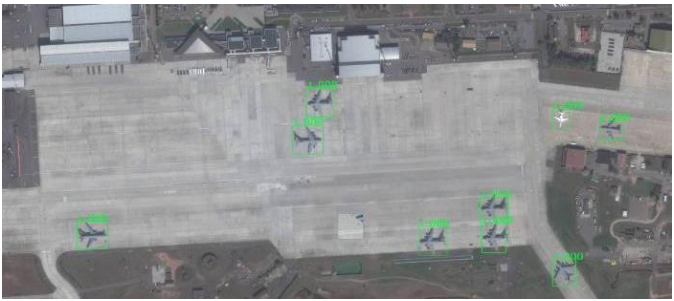

Fig. 7. Aircraft detection results (green boxes) at a military airbase in Germany. (Satellite image: DigitalGlobe, Inc.)

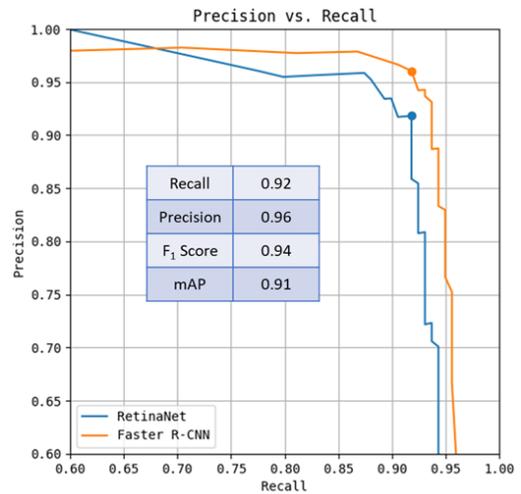

Fig. 8. Accuracy measures (for the Faster R-CNN model at the maximum $F_1$ score) and precision-recall curves of the aircraft ATR results. The two curves are for Faster R-CNN and RetinaNet. The dots mark the probability thresholds where the $F_1$ scores are maximized.

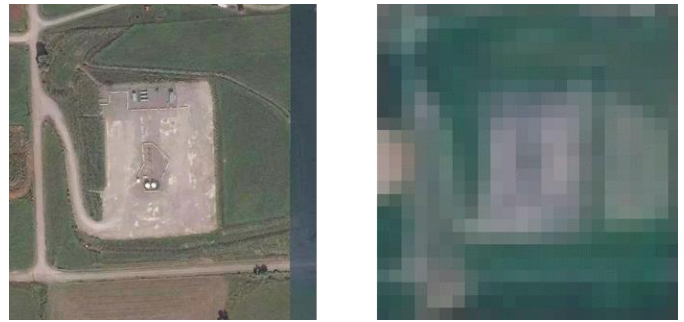

Fig. 9. Fracking well pad in WorldView-3 (left) and Sentinel-2 (right). Note that the Sentinel-2 image, whose GSD is rather low at 10 meters, lacks the fine detail of the WorldView-3 image, whose GSD is much higher at 0.3 meter. (Images: DigitalGlobe, Inc.)

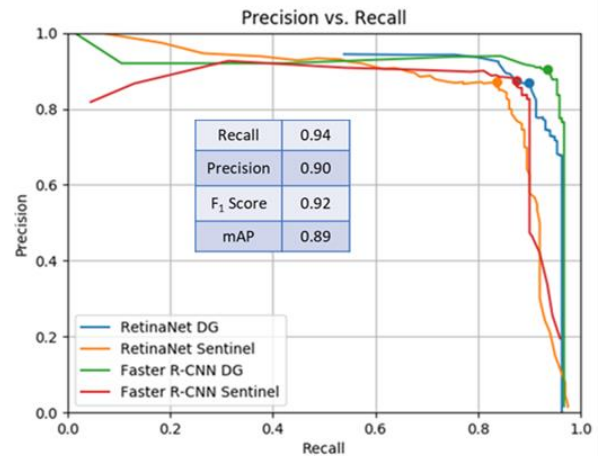

Fig. 10. Accuracy measures (for the Faster R-CNN model on WorldView-3 imagery at the maximum $F_1$ score) and precision-recall curves of the fracking well results. The four curves are for two different ATR algorithms (Faster R-CNN and RetinaNet) on two sources of satellite imagery: high-resolution WorldView-3 (labeled "DG") and low-resolution Sentinel-2. The dots on the curves mark the probability thresholds at which the $F_1$ scores are maximized.

## IV. GLOBAL SCALABILITY

To demonstrate global scalability, we trained the Faster R-CNN model to recognize fracking wells from multiple geographic regions. We curated a labeled training dataset of wells from four diverse regions associated with shale rock formations in Pennsylvania, Canada, New Mexico, and Russia. Fracking well permits for the North American locations were retrieved from state government web sites, and their latitude and longitude coordinates were downloaded to serve as initial guesses for the locations of fracking wells. WorldView-3 images at these locations were downloaded from the GBDX image server, divided into 5000x5000-pixel tiles, and manually curated at a GSD of 0.3 meter. The area of each tile was 1.5x 1.5 km or 2.25 km$^2$. For Russia, we could find no records of fracking wells on the internet. After using Google Earth to search oil fields in Siberia, we found what appeared to be fracking wells and downloaded WorldView-3 imagery from the GBDX server for manual curation as described above.

Fig. 11 shows a map of the four regions and examples of the labeled training data from each region. Note the differences in the appearances of the wells. For example, Russian well pads are larger and more irregular than those of North America. Their access roads are also much wider. The wells of New Mexico are in the desert and exhibit much less seasonal variation than the other three regions. (For more information on fracking wells, see Refs. [10] and [17].) A total of 12,214 wells from 4519 WorldView-3 images were identified and labeled. This required a total of 404 hours of manual labor. Table I summarizes the data curation.

After curation, we down-sampled the images by a factor of four to a size of 1250x1250 pixels and a GSD of 1.2 meters. We found that this reduced the inference execution time without significantly degrading the accuracy measures. We divided the training dataset into three subsets: training, validation, and test. The training dataset was used to train the ATR algorithm, and the validation dataset was used during training to ensure that the algorithm did not overfit to the training data. (Overfitting was avoided by monitoring the validation loss and the mAP on the validation data: when they plateaued, the training was halted.) After the training process finished, accuracy measures were computed from the test dataset. To avoid human bias in the selection of the test images, the software performed stratified sampling on the geographic regions, using a 70-20-10 split: 70% for training, 20% for validation, and 10% for testing. Thus, a random sample of 10% of the images from each region was set aside and reserved for measuring the accuracy. Because of the large number of images in each region, the test set had a good mix of conditions in each region (e.g., snow cover and other seasonal variations). The test set comprised a total of 1312 wells taken from 452 images, which covered an area of 1017 km$^2$.

The precision-recall curve for the global results (i.e., all four geographic regions) is shown in Fig. 12. The recall, precision, $F_1$ score, and mAP are shown in the table in the figure. The dashed light-blue line indicates a recall of 0.85. The dashed black curve indicates an $F_1$ score of 0.80. The red star on the red curve indicates the maximum $F_1$ score at which the accuracy measures are computed. We found that relaxing the localization constraint by decreasing the IoU from the conventional value of 0.50 to 0.20 improved the mAP.

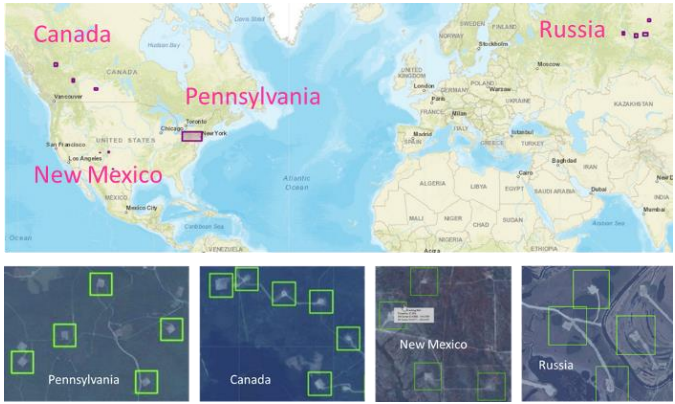

Fig. 11. Fracking wells were curated from Pennsylvania, Canada, New Mexico, and Russia. Examples of the curated images from the four regions are shown. (Satellite images: DigitalGlobe, Inc.)

TABLE I. GLOBAL DATA CURATION

| Dataset | Number of Images | Number of Wells | Curation Time (hrs) |
|---|---|---|---|
| Pennsylvania | 1031 | 1521 | 110 |
| New Mexico | 463 | 2801 | 62 |
| Canada | 1603 | 5986 | 82 |
| Russia | 1422 | 1906 | 150 |
| **Total** | **4519** | **12,214** | **404** |

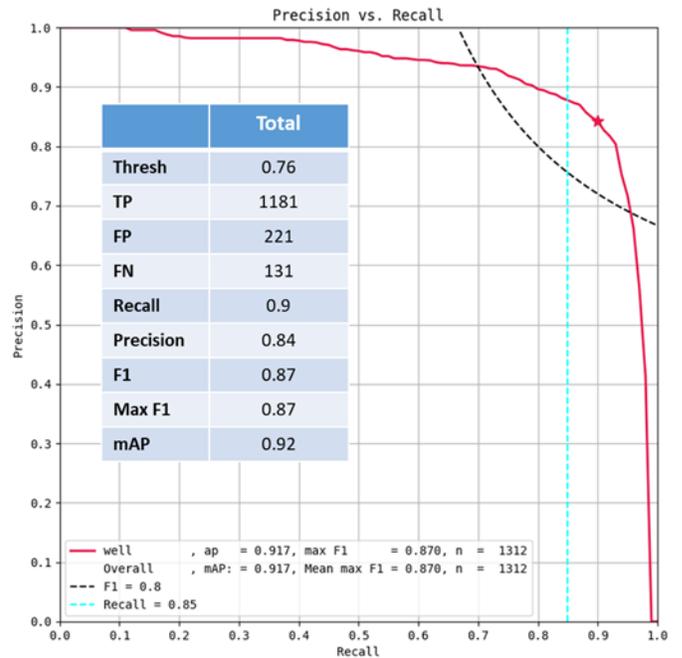

Fig. 12. Accuracy measures (at the maximum $F_1$ score) and precision-recall curve of the global results. The recall is 0.90, the precision 0.84, the $F_1$ score 0.87, and the mAP score 0.92. The star on the precision-recall curve marks the probability threshold at which the $F_1$ score is maximized.

As can be seen, the maximum $F_1$ score is 0.87. At this detection threshold, the recall is 0.90 and the precision is 0.84. The mAP is 0.92. When the software is in operation, the user can tune the detection threshold to increase or decrease the number of detections. If a high recall is desired with little regard for false detections, the threshold can be decreased. Conversely, if a low false alarm rate is desired with less regard for low recall, the threshold can be increased. The software sets the default threshold to the maximum $F_1$ score, which is marked by the star on the precision-recall curve.

How does GATR perform on a region of the world it has never seen before? We performed a "blind test" by running the Faster R-CNN model on a set of 54 images from North Dakota. The dataset covers an area of 120 km$^2$ and contains 44 wells. The region is marked on the map in Fig. 13 below, and the images show examples of the wells.

The accuracy measures on North Dakota are comparable to the tested accuracy measures shown in Fig. 12. The recall is 0.90, and the precision is 0.85. The $F_1$ score and the mAP are 0.88. The precision-recall curve is shown in Fig. 14.

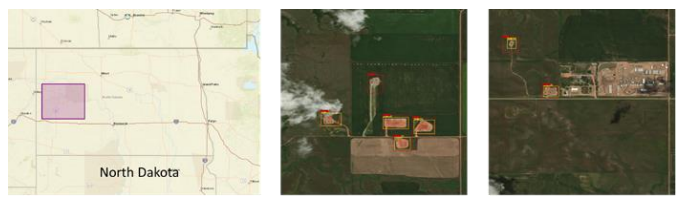

Fig. 13. Region of North Dakota where GATR was blind-tested. The images show fracking wells in this region. (Satellite images: DigitalGlobe, Inc.)

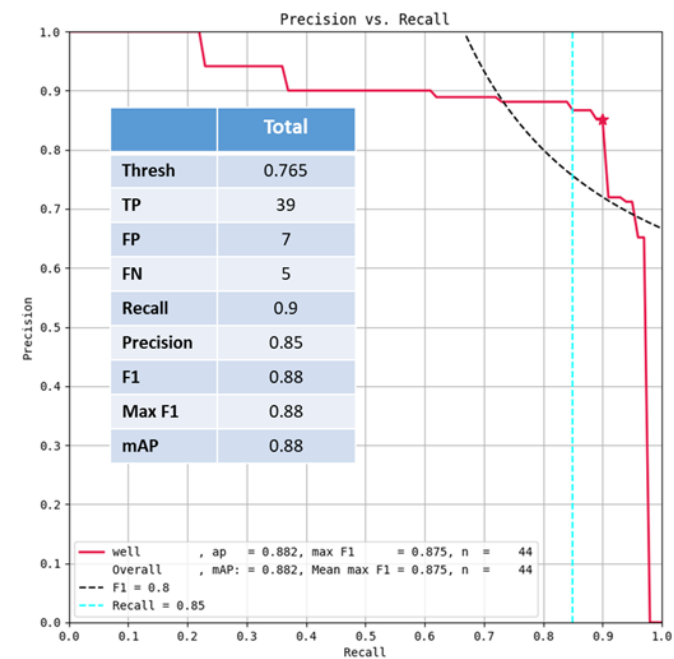

Fig. 14. Accuracy measures and precision-recall curve of the blind test on North Dakota fracking wells. The recall is 0.90, the precision is 0.85, the $F_1$ score is 0.88, and the mAP is 0.88. The star on the precision-recall curve marks the probability threshold at which the $F_1$ score is maximized. Compare these results with those of Fig. 12.

## V. SOFTWARE OPERATION

In this section we describe the operation of the GATR software. Fig. 15 shows the graphical user interface, which uses a tile map service from ESRI to provide world map data [22]. The user can pan and zoom to any region of the globe, draw a box around the area of interest, define an optional date and time constraint, specify an object type, and then search for these objects in WorldView-3 or Sentinel-2 imagery from the GBDX platform. The third image shows the results of a search for fracking wells in Pennsylvania, where the detections are shown as small green boxes. The user can zoom in for a detailed look at any of these detections, as shown in the last image. GATR saves the detections in a database along with metadata such as object type, date, time, sensor, and location.

Fig. 16 shows the results of other searches. The top image shows the results of a broad area search of the entire state of Pennsylvania. On a single NVIDIA Titan Xp GPU, it takes only two hours to search 119,000 km$^2$ and discover 3200 fracking wells. This is a processing speed of 16.8 km$^2$/sec (or more than 10 million pixels/sec). With two GPUs, the speed doubles to 35.3 km$^2$/sec, and with four GPUs, it quadruples to 68.8 km$^2$/sec. The other images show the detections of fracking wells in Canada, New Mexico, and Russia.

Fig. 17 shows a broad area search for aircraft. The first image shows a search box that is drawn over the Crimean Peninsula, which takes 12 hours to search. (The search rate is slower because of the higher image resolution needed to detect aircraft versus the lower resolution for fracking wells.) The user can zoom to any of the green detection boxes. The second and third images zoom in on a cluster of boxes at a Russian airbase, showing the detections of helicopters and aircraft. The fourth image zooms in on an isolated set of boxes that reveal an unexpected find: crop duster aircraft.

Broad area search is useful for searching large geographic regions and detecting unexpected objects, such as crop duster airplanes in Crimea. GATR can also set up "watch boxes" for monitoring sites of interest. This is useful for indications and warnings (I&W), such as alerting the user to the arrival of ships at a port, detecting the departure of aircraft from an airport, or counting objects over time.

## VI. CONCLUSION

In this paper we presented GATR, which is a modular, cloud-based software system for real-time object detection and classification in satellite imagery on a worldwide basis. GATR uses GPU-accelerated deep learning software to quickly search large geographic regions. The search time scales linearly with geographic area, and the search speed scales linearly with the number of GPUs. ATR is performed by accurate deep learning models, including RetinaNet and Faster R-CNN, which exhibit recalls of over 90%. Applications include broad area search, watch boxes for monitoring ports and airfields, and target characterization. GATR is extensible to new target types and handles radar and multispectral imagery. Currently we are developing a process of continual improvement in which feedback from analysts is incorporated in the automated retraining of the ATR algorithm. This online training function (Fig. 18) will be described in a future paper.

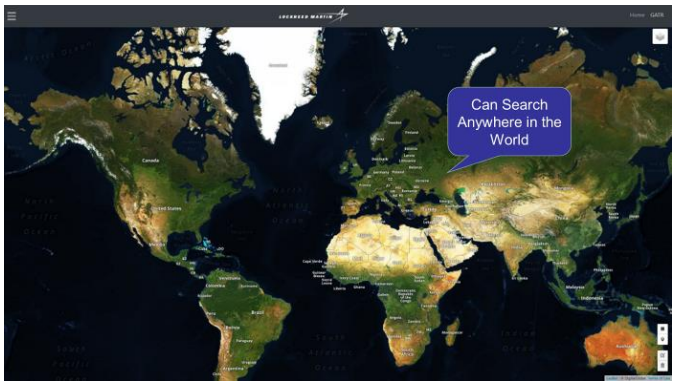
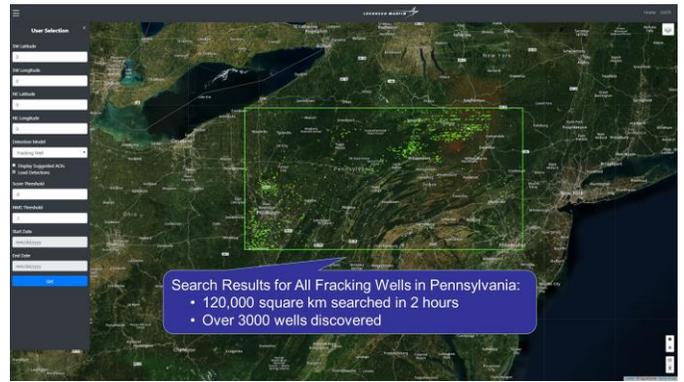
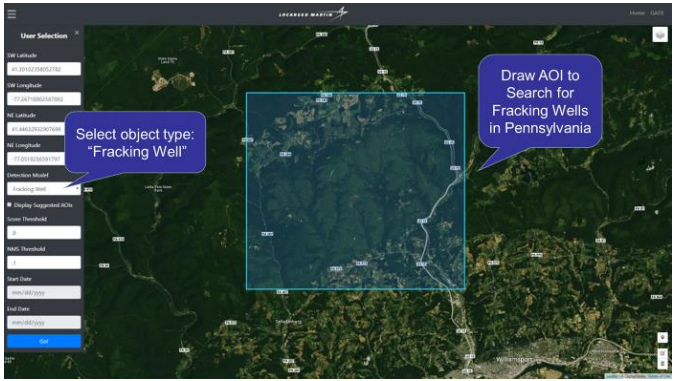
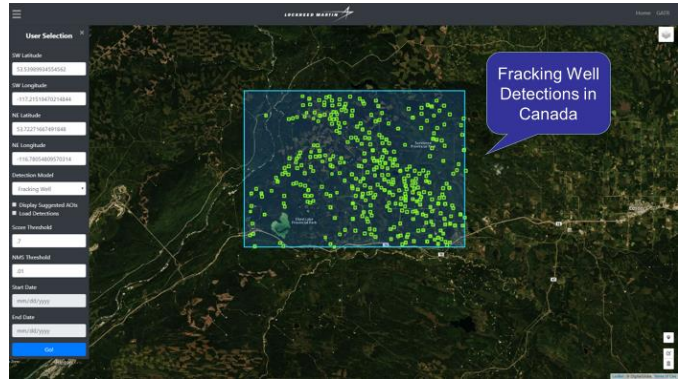
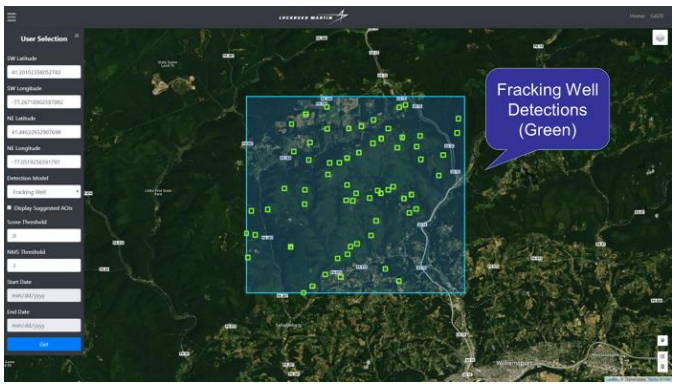
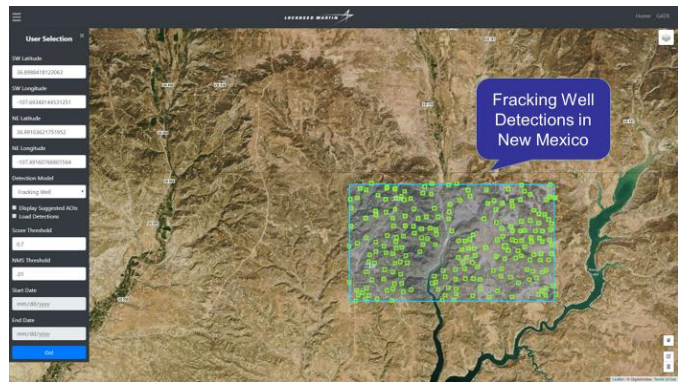
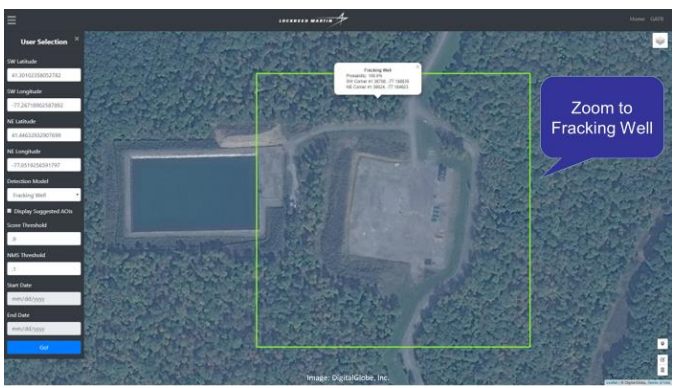
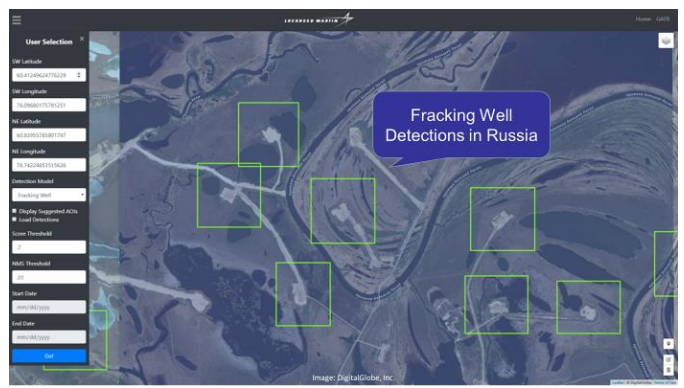

Fig. 15. Detection of fracking wells. The top image shows the GATR world map, and the second image shows the view after zooming to a region of Pennsylvania. After selecting fracking wells as the object type, pressing the "go" button, and waiting a few seconds, the user sees the detected results show up as green boxes (third image). The user can zoom to any of the detections for a closer look, as shown in the fourth image.

Fig. 16. Broad area search and global scalability. The top image shows the results of searching the entire state of Pennsylvania for fracking wells. The search of this 119,000 km$^2$ region took only two hours and detected 3200 wells. The other images show the detections of fracking wells in Canada, New Mexico, and Russia. The detections are marked as small green boxes, and the view of Russia in the last image is zoomed in to show detail.

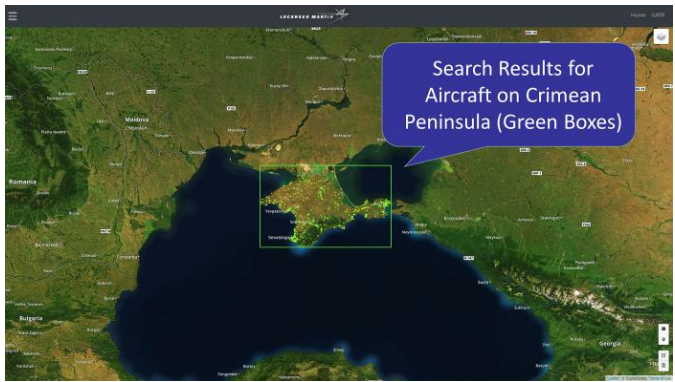

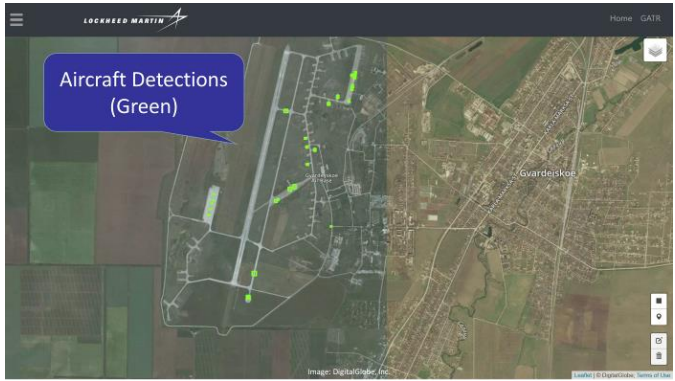

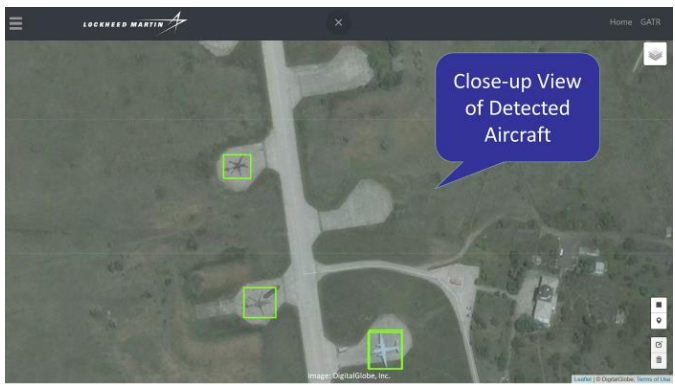

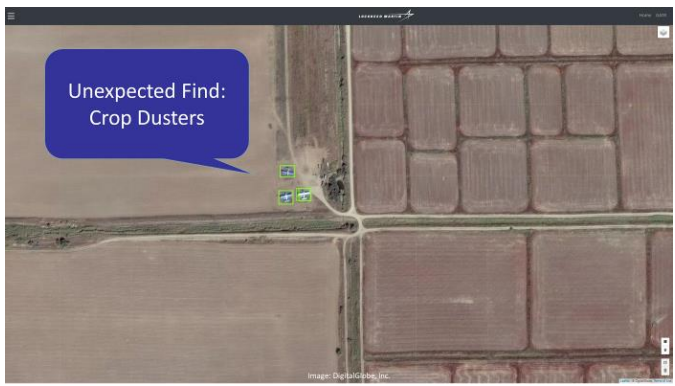

Fig. 17. Detection of aircraft. The top image shows the search box drawn on the Crimean peninsula. It takes 12 hours to search this large region in high-resolution imagery. The detections are shown as green boxes. The second and third images show zoomed-in views of a cluster of green boxes at a Russian airbase. The last image shows a zoomed-in view of other, isolated detections. They turn out to be crop duster aircraft, an unexpected find.

## References

[1] B. Schachter, Automatic Target Recognition, 3rd ed, SPIE, 2018.

[2] B. Bhanu, "Automatic target recognition: state of the art survey," IEEE Trans. Aerospace and Electronic Systems, vol AES-22, no 4, July 1986.

[3] D. Dudgeon and R. Lacoss, "An overview of automatic target recognition," Lincoln Laboratory Journal, vol 6, no 1, 1993.

[4] J. Ratches, "Review of current aided/automatic target acquisition technology for military target acquisition tasks," SPIE Optical Engineering, vol 50, no 7, July 2011.

[5] T. Lillesand, R. Kiefer, and J. Chipman, Remote Sensing and Image Interpretation, 6th ed, Hoboken: John Wiley & Sons, 2008, ch 4.

[6] "Amazon Web Services (AWS) - Cloud Computing Services", Amazon, https://aws.amazon.com.

[7] "GBDX", Maxar Technologies, https://www.geobigdata.io.

[8] "WorldView-3", DigitalGlobe, http://worldview3.digitalglobe.com/.

[9] "Sentinel-2", European Space Agency, https://sentinel.esa.int/web/sentinel/missions/sentinel-2.

[10] M. Harner, A. Groener, and M. Pritt, "Detecting the presence of vehicles and equipment in SAR imagery using image texture features," IEEE Workshop Applied Imagery Pattern Recognition (AIPR), Oct 2019.

[11] "Sentinel-1", European Space Agency, https://sentinel.esa.int/web/sentinel/missions/sentinel-1.

[12] M. Pritt and G. Chern, "Satellite image classification with deep learning," IEEE Workshop Applied Imagery Pattern Recognition (AIPR), Oct 2017.

[13] M. Pritt, "Deep learning for recognizing mobile targets in satellite imagery," IEEE Workshop Applied Imagery Pattern Recognition (AIPR), Oct 2018.

[14] "TensorFlow: An open-source software library for machine intelligence," TensorFlow, https://www.tensorflow.org.

[15] "PyTorch", PyTorch, https://pytorch.org.

[16] F. Chollet, "Keras", GitHub, 2017, https://github.com/fchollet/keras.

[17] A. Groener, G. Chern, and M. Pritt, "A comparison of deep learning object detection models for satellite imagery," IEEE Workshop Applied Imagery Pattern Recognition (AIPR), Oct 2019.

[18] T.-Y. Lin et al., "Focal loss for dense object detection," arXiv 1708.02002, 7 Feb 2018.

[19] S. Ren, K. He, R. Girshick and J. Sun, "Faster R-CNN: towards real-time object detection with region proposal networks," arXiv 1506.01497v3 [cs.CV], 6 Jan 2016.

[20] "DIU xView 2018 Detection Challenge," DIU, http://xviewdataset.org.

[21] D. Lam et al., "xView: objects in context in overhead imagery," arXiv 1802.07856, 22 Feb 2018.

[22] "ArcGIS," ESRI, https://www.arcgis.com/home/index.html.

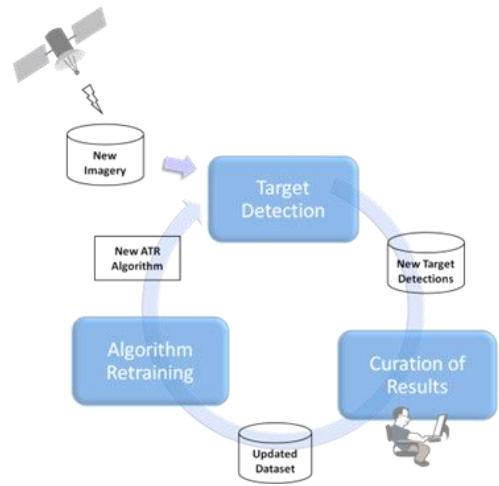

Fig. 18. Online training for continual improvement of GATR's accuracy.

7